\documentclass[pdflatex, sn-mathphys]{sn-jnl}

\usepackage{graphicx}
\usepackage{caption}
\usepackage{subcaption}

\usepackage{color, soul}

\soulregister\cite7
\soulregister\ref7
\soulregister\pageref7

\jyear{2023}%

\theoremstyle{thmstyleone}%
\theoremstyle{thmstyletwo}%

\theoremstyle{thmstylethree}%

\begin{document}

\title[Using Hand Pose Estimation To Automate Open Surgery Training Feedback]{Using Hand Pose Estimation To Automate Open Surgery Training Feedback}

\author[1]{\fnm{Eddie} \sur{Bkheet}}\email{eddie.bkheet@gmail.com}

\author[2]{\fnm{Anne-Lise} \sur{D'Angelo}}\email{annelise.dangelo@gmail.com}

\author[3]{\fnm{Adam} \sur{Goldbraikh}}\email{goldb.adam@gmail.com}

\author[1]{\fnm{Shlomi} \sur{Laufer}}\email{laufer@technion.ac.il}

\affil[1]{\orgdiv{Data and Decision Sciences}, \orgname{Technion Institute of Technology}, \orgaddress{\city{Haifa}, \country{Israel}}}

\affil[2]{\orgdiv{Surgery}, \orgname{Mayo Clinic}, \orgaddress{\city{Rochester}, \state{Minnesota}, \country{United States}}}

\affil[3]{\orgdiv{Applied Mathematics}, \orgname{Technion Institute of Technology}, \orgaddress{\city{Haifa}, \country{Israel}}}

\abstract{
\textbf{Purpose:} This research aims to facilitate the use of state-of-the-art computer vision algorithms for the automated training of surgeons and the analysis of surgical footage. By estimating 2D hand poses, we model the movement of the practitioner's hands, and their interaction with surgical instruments, to study their potential benefit for surgical training.

\textbf{Methods:} We leverage pre-trained models on a publicly-available hands dataset to create our own in-house dataset of 100 open surgery simulation videos with 2D hand poses. We also assess the ability of pose estimations to segment surgical videos into gestures and tool-usage segments and compare them to kinematic sensors and I3D features. Furthermore, we introduce \textbf{6} novel surgical dexterity proxies stemming from domain experts' training advice, all of which our framework can automatically detect given raw video footage.

\textbf{Results:} State-of-the-art gesture segmentation accuracy of 88.35\% on the Open Surgery Simulation dataset is achieved with the fusion of 2D poses and I3D features from multiple angles. The introduced surgical skill proxies presented significant differences for novices compared to experts and produced actionable feedback for improvement. 

\textbf{Conclusion:} This research demonstrates the benefit of pose estimations for open surgery by analyzing their effectiveness in gesture segmentation and skill assessment. Gesture segmentation using pose estimations achieved comparable results to physical sensors while being remote and markerless. Surgical dexterity proxies that rely on pose estimation proved they can be used to work towards automated training feedback. We hope our findings encourage additional collaboration on novel skill proxies to make surgical training more efficient.
}




\maketitle

\section{Introduction}\label{intro}

Until this day, the traditional surgical training methodology of “see one, do one, teach one” by Dr. William Halsted \cite{kotsis_application_2013} remains the most practiced method among medical practitioners. A major drawback of this methodology is the limited availability of expert surgeons. To make the training process more efficient, several works have been done that aim to recognize gestures and estimate surgical skill levels using different input modalities, including kinematic sensory data \cite{fawaz_evaluating_2018, goldbraikh_using_2022}, optical flow \cite{sarikaya_surgical_2020}, and RGB videos \cite{funke_video-based_2019, goldbraikh_video-based_2022, lavanchy_automation_2021, liu_towards_2021, wang_towards_2021, zhang_sd-net_2021}.

While sensor-based models \cite{fawaz_evaluating_2018, goldbraikh_using_2022} provide accurate spatial coordinates of the hands and surgical instruments, they can be challenging to implement due to the expensive hardware and setup required. In contrast, computer vision models that utilize RGB videos as input present a more viable solution, as they are less intrusive to the surgical workflow and are easier to set up.

Some of the popular datasets in this field are the JIGSAWS dataset \cite{gao_jhu-isi_nodate} and the more recent RARP-45 dataset \cite{van_amsterdam_gesture_2022}. Existing studies focus on hand detection \cite{zhang_using_2020, louis_temporally_2021, goodman_real-time_2021}, surgical instrument detection \cite{jin_tool_2018, goldbraikh_video-based_2022}, tool usage \cite{basiev_open_2022, goldbraikh_video-based_2022}, surgical gesture recognition \cite{goldbraikh_using_2022, zhang_sd-net_2021, sarikaya_surgical_2020, funke_video-based_2019}, and surgical skill classification \cite{fawaz_evaluating_2018, lavanchy_automation_2021, liu_towards_2021, wang_towards_2021, zhang_sd-net_2021}. Previous studies \cite{gesture_keypoints_warping, goodman_real-time_2021} have also demonstrated the effectiveness of pose estimations for the task of gesture recognition.

Existing work on skill assessment focuses on classifying the performance using categorical labels or a numeric score and statistically comparing the performance of novices to experts. Existing frameworks offer feedback that includes (1) highlighting the segments that contributed to the skill classification \cite{fawaz_evaluating_2018} and (2) providing \textbf{global} motion metrics and statistics for novice compared to experts \cite{goldbraikh_using_2022, lavanchy_automation_2021, goldbraikh_video-based_2022}. To the best of our knowledge, a fully automated framework that performs explainable skill assessment with task-specific actionable feedback hasn't been released yet.

This research aims to bridge the gap in the implementation of computer vision models in the surgical training environment by exploring surgical dexterity proxies based on 2D hand pose estimations, that automate the expert's advice. The proposed proxies produce \textbf{task-specific} feedback focused on surgical tool utilization for scissors, needle drivers, and forceps, as well as suture holding, and cutting positions. 

This paper’s contributions are as follows:
\begin{enumerate}
  \item A fully automated modular surgical dexterity assessment framework with \textbf{actionable} and explainable feedback.
  \item A novel 2D hand pose dataset of open surgery simulation with annotations of instruments, hands, tool usage, gestures, and skill levels.
  \item State-of-the-art results on the multi-task gesture segmentation problem using the predicted skeletons.
\end{enumerate}

\section{The Dataset}
\label{dataset}
The dataset used in this research is the Open Surgery Simulation dataset \cite{goldbraikh_using_2022}. It contains 100 videos of 25 clinicians with varying levels of expertise. The dataset contains temporal segmentation annotations for tool utilization and gesture recognition. The tools are needle driver, scissors, and forceps. The gestures are "No Gesture", "Needle Passing", "Pull The Suture", "Instrumental Tie", "Lay The Knot", and "Cut The Suture". The dataset was split into 5 folds using cross-user validation (Leave One User Out \cite{gao_jhu-isi_nodate} modified for groups of users).

In order to evaluate the added value of pose estimations to surgical skill assessment, 925 frames (0.01\% of the dataset) sampled from 15 videos were annotated with 2D hand poses that include bounding boxes and 21 key-points. This results in a total of 1987 annotated hand instances.

\section{Methods}
\label{methods}

\subsection{Object Detection}
\label{methods:1}
For the task of hands and tools detection, a real-time YOLO-X \cite{ge_yolox_2021} object detection model was used. The model configuration was according to the YOLOX-S version as in the official release. The model was trained using the "mmdetection" framework \cite{chen_mmdetection_2019} for 400 epochs, each epoch took about 79 seconds on our hardware.
The weights were initialized using a pre-trained model on the COCO \cite{lin_microsoft_2015} dataset. The prediction head was modified to detect 2 hands (left and right) and 4 surgical tools (needle driver, scissors, forceps, forceps-not-used). The model was evaluated using the mean average precision (mAP) of intersection over union (IoU). Finally, the trained model was used to extract per-frame bounding boxes using a confidence threshold of 0.5 where a single bounding box with the highest confidence threshold was kept for each class. The model runs inference at 38 FPS on our hardware.

\subsection{2D Pose Estimation}
\label{methods:2}
The pose estimation model was trained using the "mmpose" framework \cite{mmpose2020}. HRNetV2-W18 \cite{sun_deep_2019} was compared to Simple-Baseline \cite{haiping_simple_2018} with Resnet 50 backbone, and the latter was used during the succeeding stages due to its speed of 43 items/second compared to HRNet's 9 items/second on our hardware. These models were evaluated using the probability of correct keypoint (PCK), area under the curve (AUC), and endpoint error (EPE) metrics. The models were pre-trained on the OneHand \cite{wang_mask-pose_2019} dataset and fine-tuned on our dataset for 85 additional epochs. The trained model was used to extract per-frame pose estimations using a key point confidence threshold of 0.3. The missing keypoints were imputed using the last observation carried forward (LOCF) method. Finally, the Savitzky-Golay \cite{savitzky_smoothing_1964} signal smoothing algorithm was used to minimize the jittering of key-points across time caused by the frame-wise inference.

\begin{figure}[ht]
\begin{subfigure}{1.0\textwidth}
  \centering
  \includegraphics[width=\textwidth]{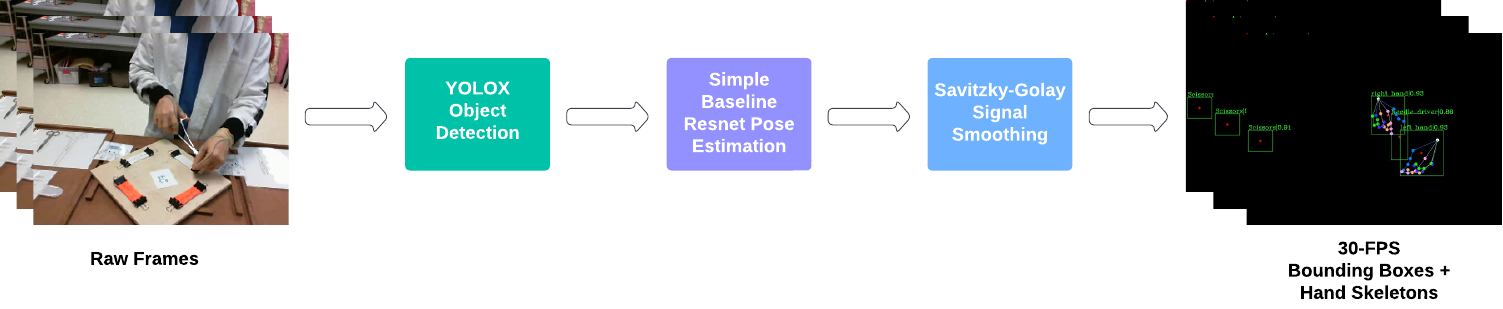}  
\end{subfigure}
\begin{subfigure}{1.0\textwidth}
  \centering
  \includegraphics[width=\textwidth]{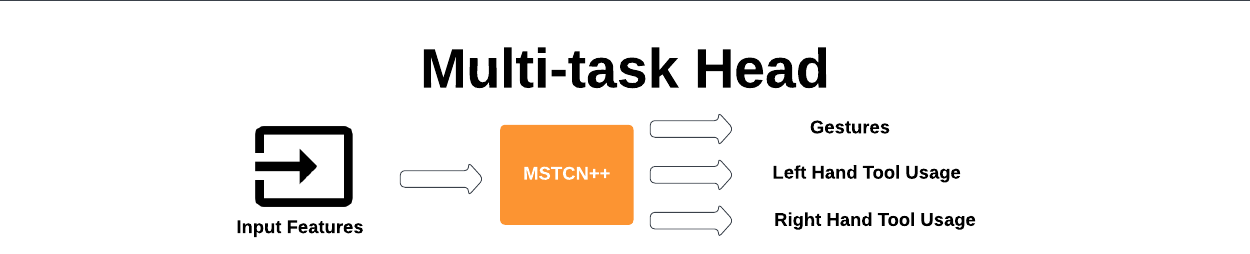} 
\end{subfigure}
\caption{Data Pipeline}
\label{fig:datapipeline}
\end{figure}

\subsection{Multi-task Temporal Activity Segmentation}
\label{methods:4}
The extracted keypoints, and the bounding box centers are concatenated to form the \textbf{input features} \label{detectionsinput} of our spatio-temporal deep learning architecture as seen in Fig. \ref{fig:datapipeline}. The architecture uses MSTCN++ \cite{li_ms-tcn_2020} to predict the performed gesture and tools utilized for each frame by modifying the prediction head to produce multiple predictions. Our input features were compared to I3D \cite{carreira_quo_2018} as a baseline feature extraction method for video processing, and to the combination of our input and I3D features. The I3D model was pre-trained on kinetics-400 \cite{kay_kinetics_2017} and fine-tuned on our dataset for 100 epochs taking 265 seconds per epoch. The features were extracted with a sliding window of 32 frames and a stride of 16 at 127 FPS. Our results were also compared to the current state-of-the-art benchmark on this dataset which uses kinematic sensor data \cite{goldbraikh_using_2022}. The hyper-parameter space of MSTCN++ was explored using the random search strategy with 100 trials. The range for learning rate was [0.0001, 0.0005, 0.005, 0.001], for feature maps it was [32, 64, 128, 256], for prediction generation layers it was [3, 7, 11, 15], for refinement layers it was [6, 10, 14], and for refinement stages it was [1, 3, 6, 12].

\begin{figure}[ht]
\begin{minipage}[t]{.45\textwidth}
\centering
\includegraphics[width=1\textwidth, height=5cm]{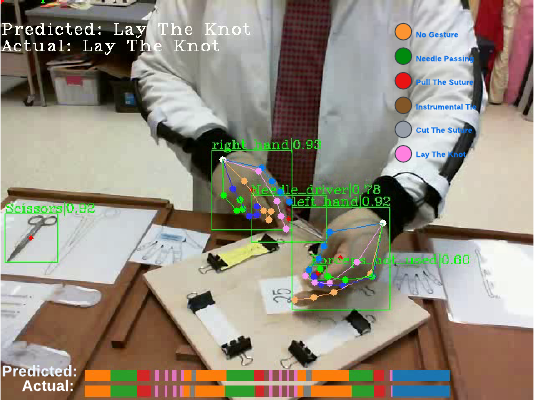}

\caption{Gesture Segmentation on Test Sample Frontal View}
\label{fig:gesture-segmentation-bars-test}
\end{minipage}
\hfill
\begin{minipage}[t]{.45\textwidth}
\centering
\includegraphics[width=1\textwidth, height=5cm]{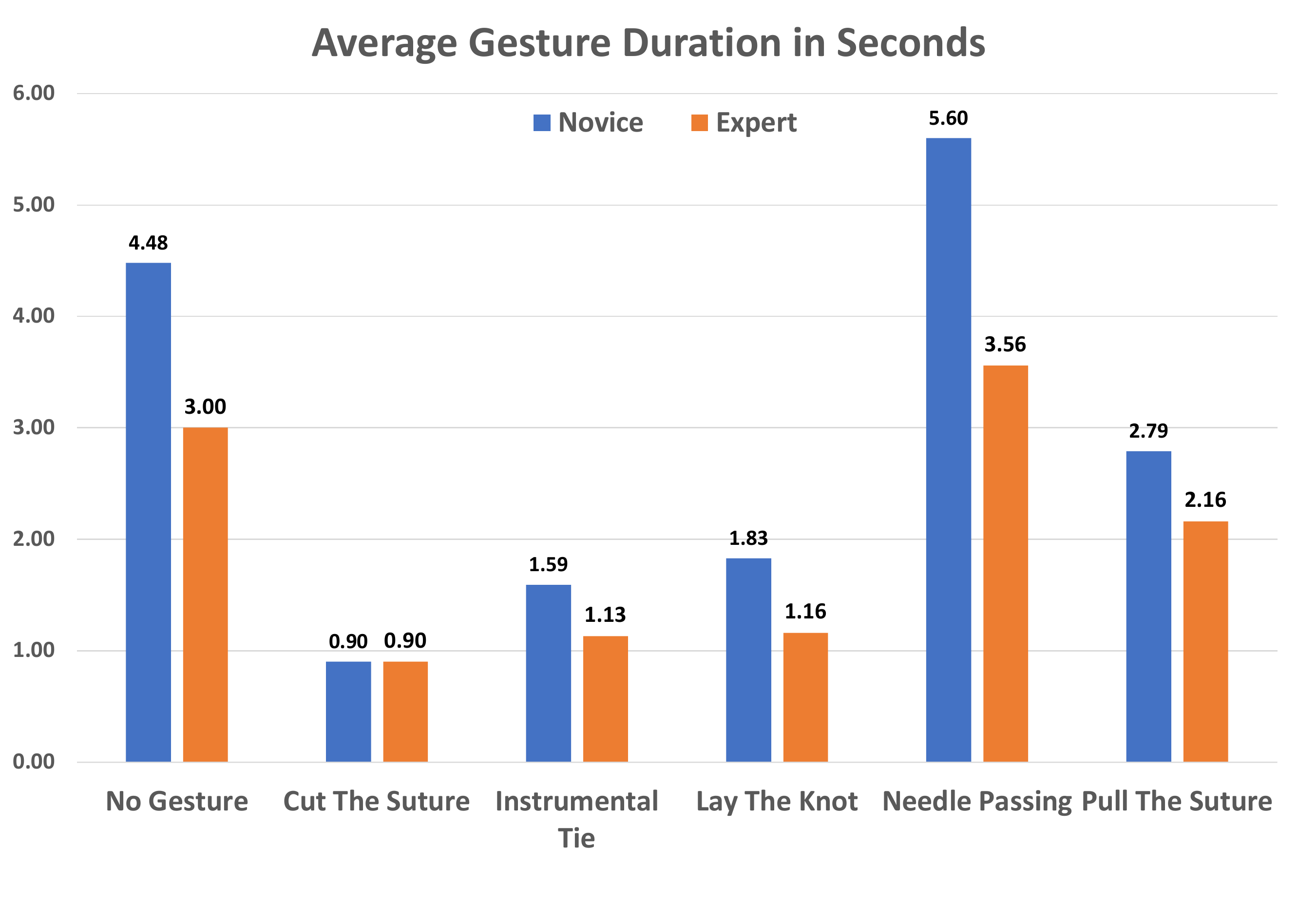}
\caption{Duration Per Gesture In Seconds}
\label{fig:durationpergesture}
\end{minipage}
\end{figure}

\subsection{Surgical Skill Assessment}
\label{methods:5}

For the purpose of surgical skill assessment, we propose an approach that relies on the building block of a \textit{Surgical Skill Proxy} \cite{liu_towards_2021}. In our case, a surgical skill proxy is defined as a metric that can be calculated using the tools and skeletons in the video, and that can be explained to the performer in simple words. In order to measure the statistical significance of the proxy differences between novices and experts, a 2-sided t-test was used which assumes a normal distribution of the proxy means across participants per task occurrence.

In order to come up with meaningful proxies and provide evidence of validity, a domain expert (Author AD) was consulted during the proxy engineering process to provide face validity. The initial proxies were created based on known techniques, and later on, new ones were identified based on the data in an iterative process. Following are the proxies that showed promising feedback and how they model the practitioner's performance.

\subsubsection{Gesture Duration}
\label{methods-gesture-duration:5.1}
\quad

\textbf{Proxy measurement:} This proxy measures the time spent on a particular portion of the operation.

\textbf{Clinical relevance:} Previous studies \cite{goldbraikh_using_2022} have shown differences in duration between novice and expert when performing different gestures. Furthermore, the background gesture which indicates the time spent on setup and non-specific movement between gestures tends to be longer for residents as observed in figure \ref{fig:durationpergesture}.

\subsubsection{Hand Orientation}
\label{methods-hand-orientation:5.2}
\quad

\textbf{Proxy measurement:} This proxy measures the level of pronation of the hand as seen in figure \ref{fig:proxyvisualization}. It's calculated as $x_{index}-x_{pinky}$ where $x_i$ is the x coordinate of the key point corresponding to the Metacarpophalangeal (MCP) joints of the i'th finger. The proxy produces high positive values when the hand is in a completely pronated position (palm rotated down), values around zero when the hand is directed to the sides (palm to the side and thumb up or down), and low negative values when the hand is fully supinated (palm rotated up).

\textbf{Clinical relevance:} Research has shown that pronation and supination of the hand are considered key skills when learning to perform surgical sutures. For example, when cutting the suture, full supination of the hand leads to the scissors forming a 90-degree angle with the suture, increasing the chance of cutting the knot. Therefore, slight supination of the hand is encouraged to form a 45-degree angle instead \cite{trott_2012}. When holding forceps, full pronation of the hand indicates an incorrect holding position, whereas slight supination leads to a pencil-like holding position, as is taught by surgical experts \cite{trott_2012}. 
Another example is needle passing, where the literature instructs starting with a pronated position, and ending the gesture with a slightly supinated position \cite{trott_2012}. Beginning the gesture in a supinated position in this case limits the freedom of hand movement, resulting in awkward hand positions with less granularity of control.

\subsubsection{Distance Between Thumb and Index Fingers}
\label{methods-distance-thumb-index:5.3}
\quad

\textbf{Proxy measurement:} This proxy measures the distance in pixels between the tip of the thumb and index fingers as seen in figure \ref{fig:proxyvisualization}. 

\textbf{Clinical relevance:} The basic method of holding a needle driver is to straighten the index finger while slightly inserting the thumb through one of the handles. This way, the index finger guides and stabilizes the needle driver while the thumb is used to open and close it \cite{kantor_2016}. A more advanced technique is to palm the needle driver, by resting one side of the handle on the thenar eminence rather than inserting the thumb into the handle and using the thenar eminence to open and close it. This has the advantage of allowing a wider range of rotational motion of the instrument within the hand \cite{kantor_2016}. Another example is the suture holding position. When holding the suture, we observed that experienced surgeons tend to hold the suture at the edge of their fingers, allowing them a stronger grip on the suture, whereas novice surgeons end up holding the suture with different parts of their thumb, perhaps due to the larger surface area, allowing easier yet less granular grip of the suture.

\subsubsection{Fingers to Tissue Distance}
\label{methods-distance-fingers-tissue:5.4}
\quad

\begin{figure}[ht]
\begin{subfigure}{1.0\textwidth}
  \centering
  \includegraphics[width=\textwidth]{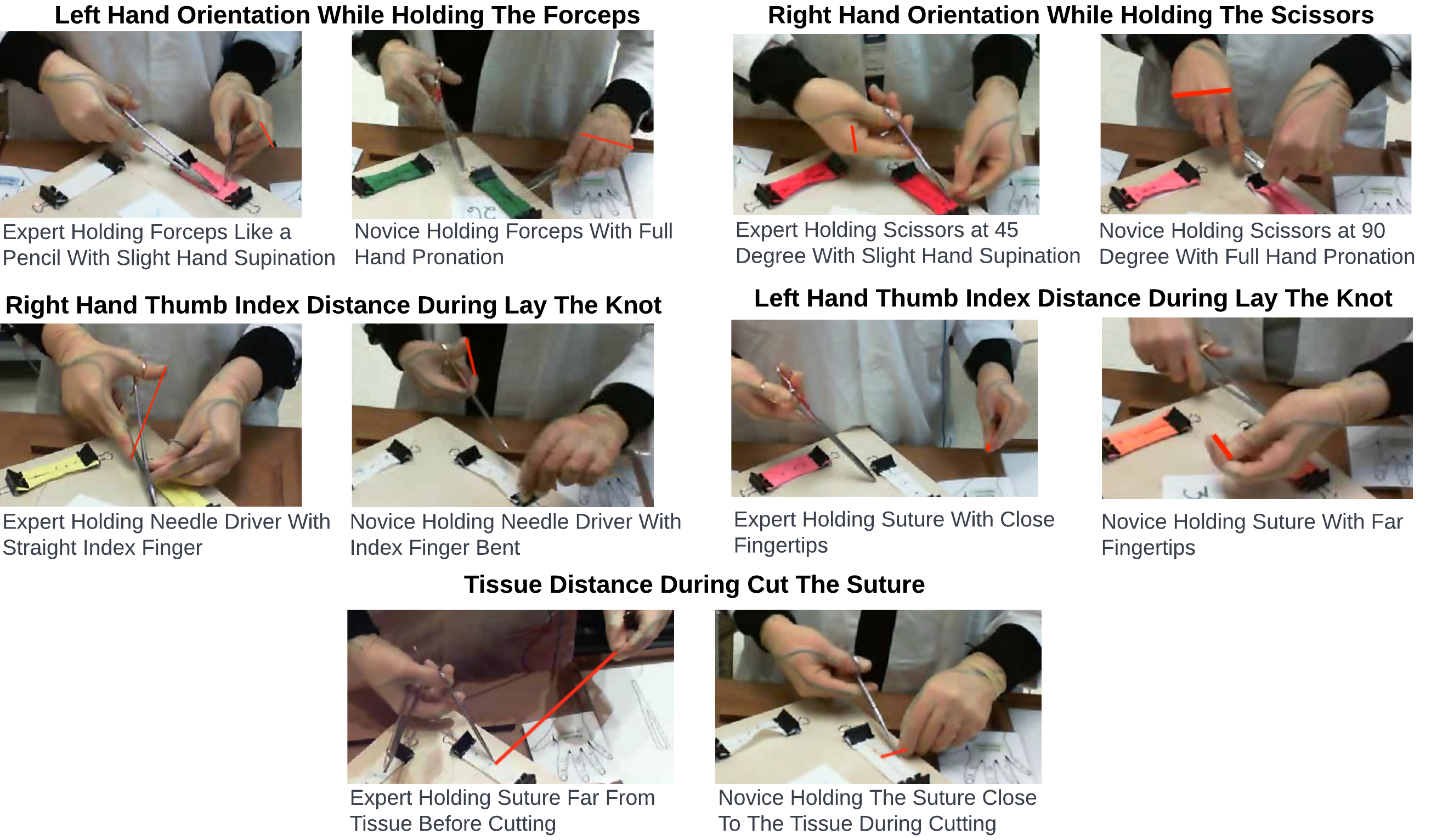} 
\end{subfigure}
    \caption{Surgical Proxy Visualization}
    \label{fig:proxyvisualization}
\end{figure}

\textbf{Proxy measurement:} This proxy measures the distance in pixels between the tissue and the fingers holding the suture as seen in figure \ref{fig:proxyvisualization}.

\textbf{Clinical relevance:} When cutting the suture, the proximity of the hand to the tissue impacts the surgeon's field of view. A small distance could lead to obscuring the suture, making the cut prone to errors. Oftentimes, assistants cut sutures for the leading surgeon, therefore, holding the suture at distance is required for better surgical performance.

\subsubsection{Hand Velocity}
\label{methods-hand-velocity:5.5}
\quad

\textbf{Proxy measurement:} This proxy measures the velocity of the hand when pulling the suture using a needle driver. The velocity of the hand is measured from multiple key points to account for subtle rotations of the hand. Such movements might be undetectable solely using object detection methods.

\textbf{Clinical relevance:} When pulling the suture, the surgeon needs to leave just the right amount of tail to allow for efficient performance during instrument ties \cite{mastenbjork_meloni_2019}. Too short of a tail does not allow enough suture to complete the knot and too long of a tail makes it cumbersome to pull through the loop to form the knot. Novice surgeons who aren't familiar with the gesture, need to move their hand slowly while keeping an eye on the tail to leave the right amount, whereas expert surgeons who are familiar with the gesture move, their hand faster consistently leaving an accurate tail length.

\section{Experiments and Results}
\label{experiments_and_results}

\begin{figure}
     \centering
     \begin{subfigure}[b]{0.15\textwidth}
         \centering
         \includegraphics[width=\textwidth,height=5cm]{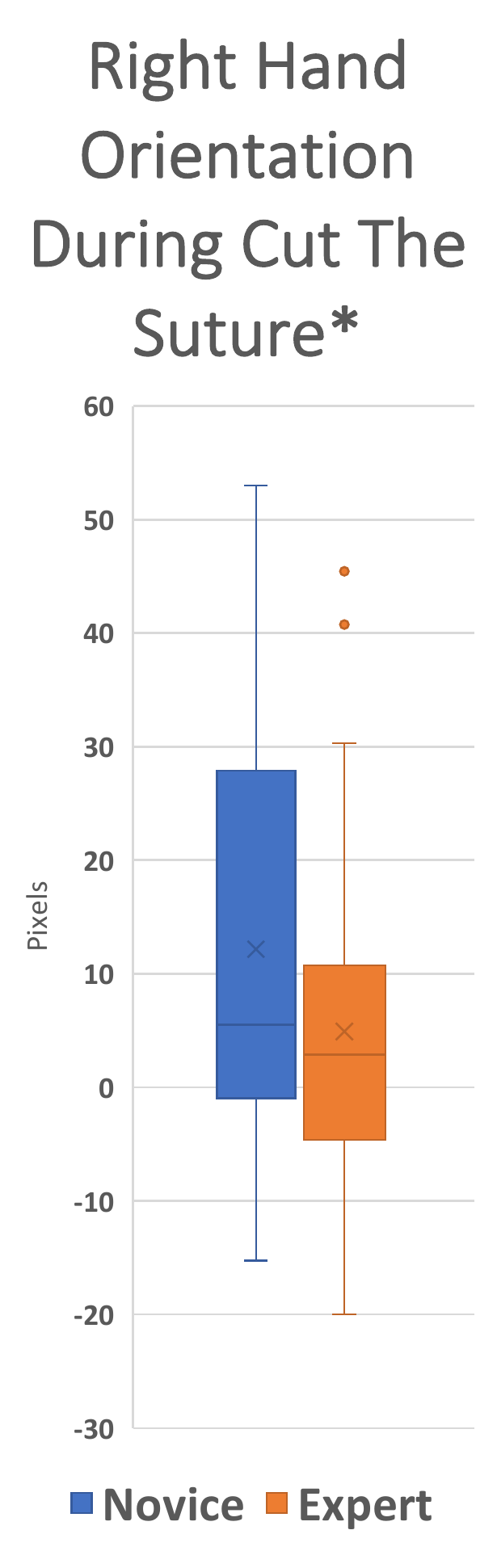}
     \end{subfigure}
     \hfill
     \begin{subfigure}[b]{0.15\textwidth}
         \centering
         \includegraphics[width=\textwidth,height=5cm]{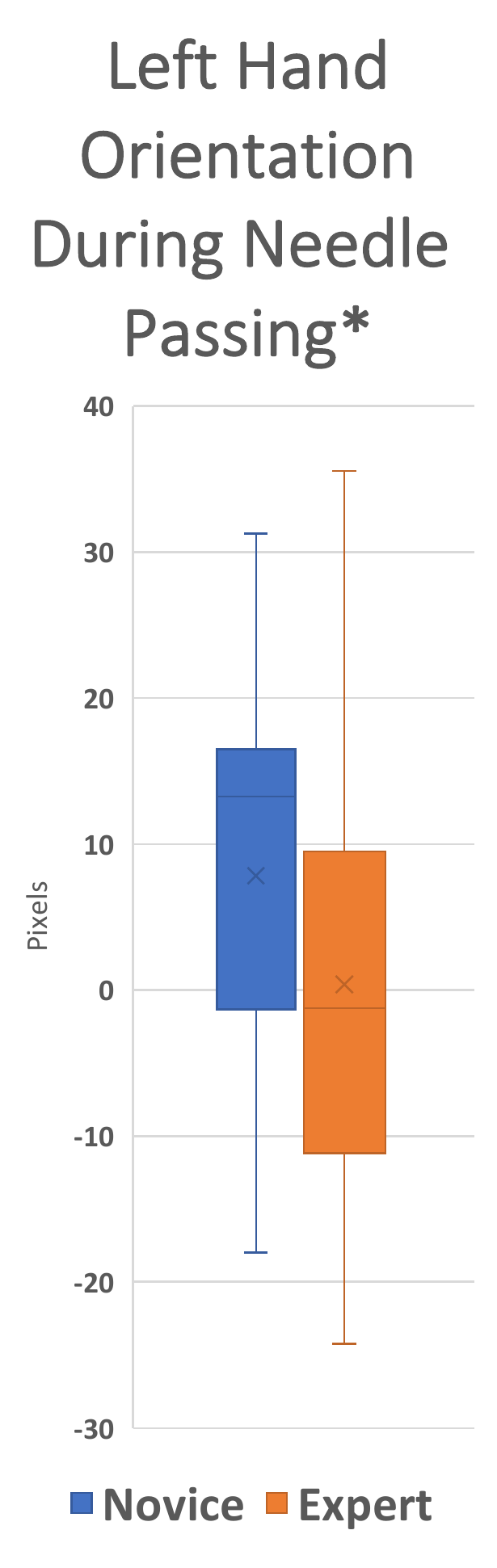}
     \end{subfigure}
     \hfill
     \begin{subfigure}[b]{0.15\textwidth}
         \centering
         \includegraphics[width=\textwidth,height=5cm]{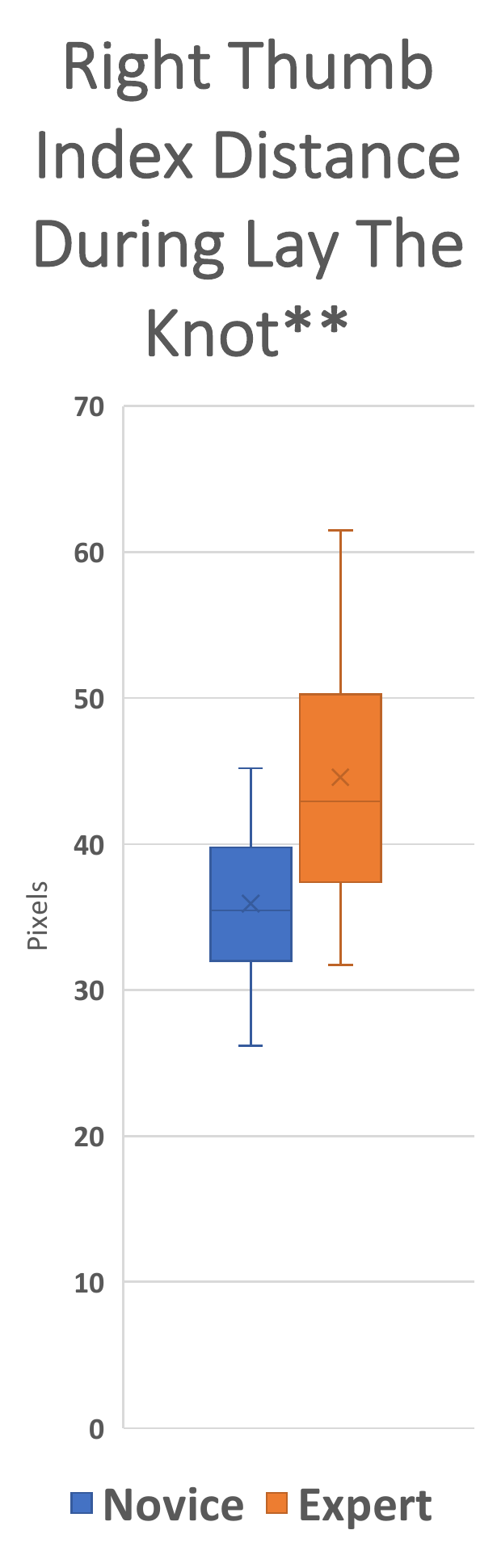}
     \end{subfigure}
     \hfill
     \begin{subfigure}[b]{0.15\textwidth}
         \centering
         \includegraphics[width=\textwidth,height=5cm]{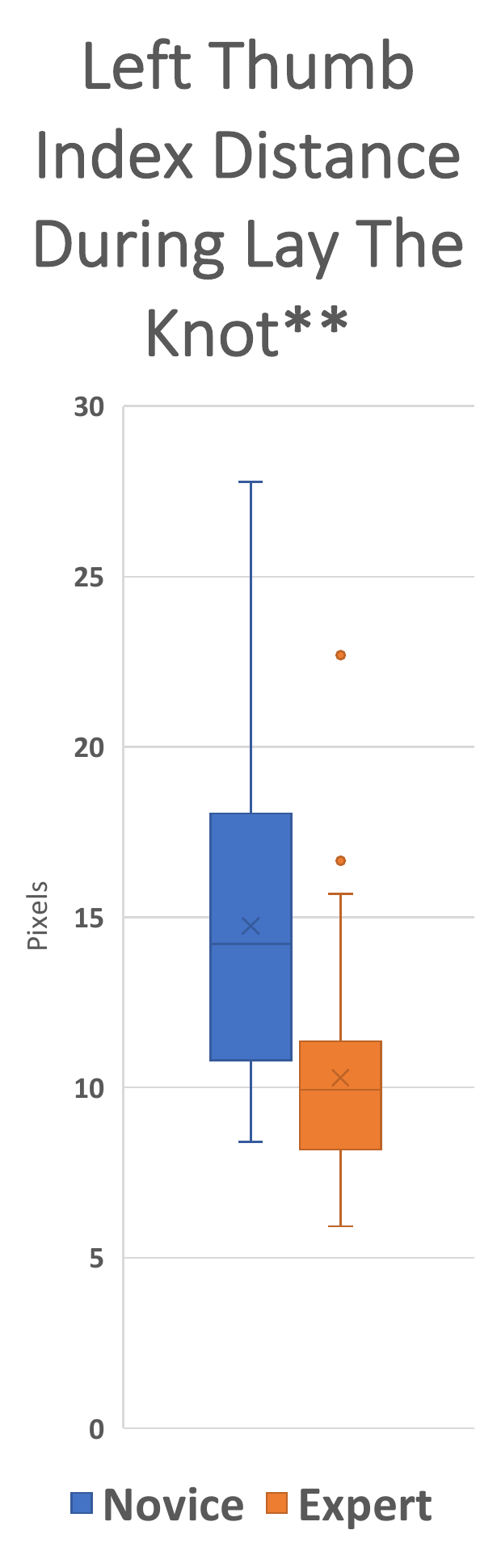}
     \end{subfigure}
     \hfill
     \begin{subfigure}[b]{0.15\textwidth}
         \centering
         \includegraphics[width=\textwidth,height=5cm]{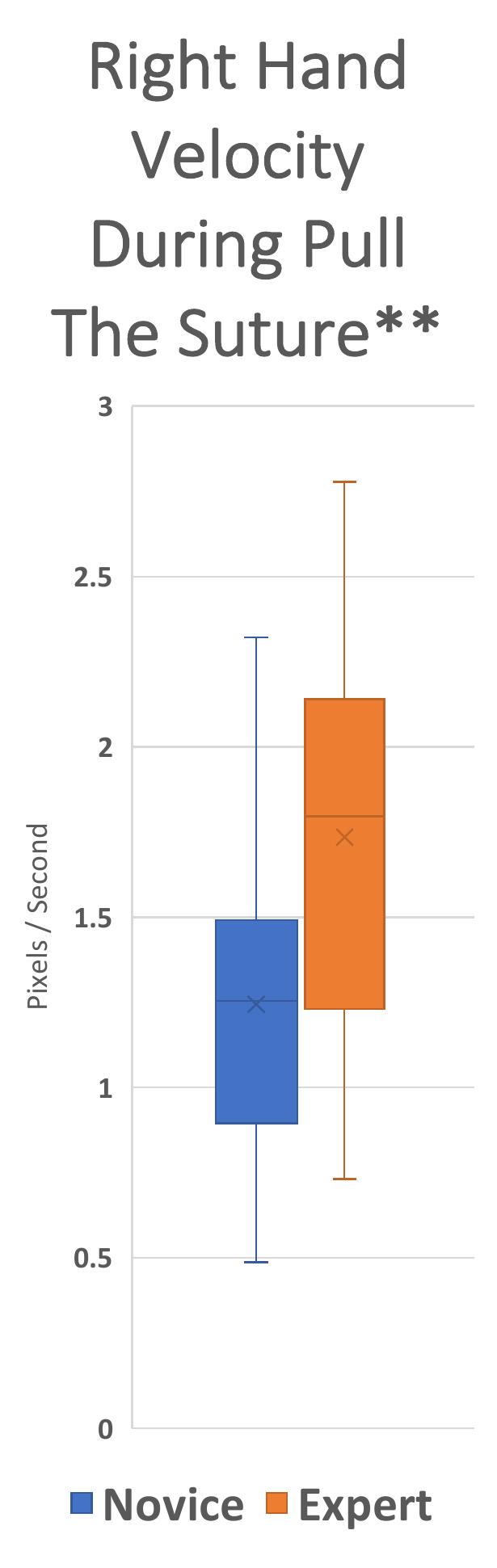}

     \end{subfigure}
     \hfill
     \begin{subfigure}[b]{0.15\textwidth}
         \centering
         \includegraphics[width=\textwidth,height=5cm]{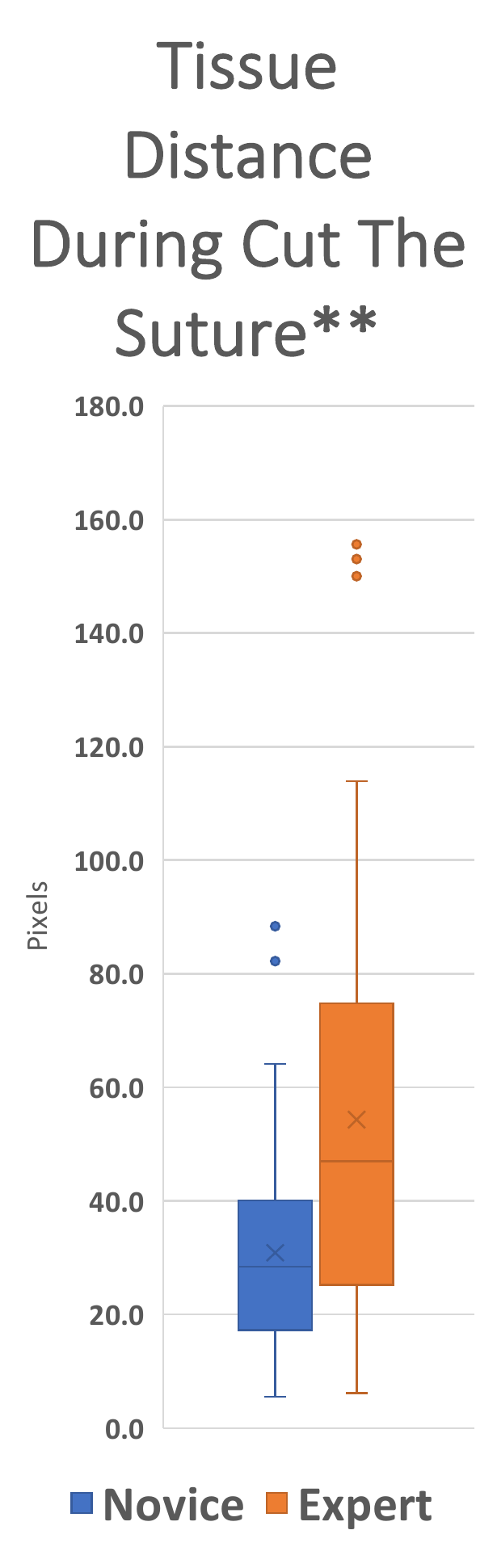}
     \end{subfigure}
     \hfill
        \caption{Surgical Dexterity Proxy Novice Vs. Expert \\(*) for p-value $<$ 0.05, (**) for p-value $<$ 0.005}
        \footnotetext{The box plots show a comparison of the mean proxy values for novices compared to experts.}
        \label{fig:proxystatistics}
\end{figure}

Table \ref{poseresults} shows the pose estimation results of HRNet compared to Simple Baseline. Simple Baseline shows comparable accuracy but offers a significant advantage in speed. The gesture segmentation results of our multi-task network are presented in table \ref{gesturesegmentationresults}. The highest accuracy of 88.35\% was achieved from the fusion of 2D poses and I3D visual features from multiple angles as seen in figure \ref{fig:gesture-segmentation-bars-test}. Figure \ref{fig:proxystatistics} shows a box plot of mean proxy values of novices compared to experts for the presented proxies during the relevant gestures.\\
All experiments were conducted on a Tesla V100 GPU with 32GB memory.

\begin{table}[ht]
\begin{center}
\begin{minipage}{\textwidth}
\caption{2D Hand Pose Estimation Results}\label{poseresults}%
\begin{tabular}{@{}llllll@{}}
\toprule
\textbf{Model Name} &  \textbf{PCK}  & \textbf{AUC} & \textbf{EPE} & \textbf{ETT\textsuperscript{1}} & \textbf{Item/s}  \\
\midrule
\textbf{Resnet Simple Baseline Pre-trained}  &	0.729 &	0.567 &	15.107 & - &	43.47\\
\textbf{HRNet Pre-trained}   &	0.771 &	0.603 &	13.279 & - &	9.52\\
\textbf{Resnet Simple Baseline Fine-tuned}  &	0.949 &	0.774 &	7.178 & 17 &	\textbf{43.47}\\
\textbf{HRNet Fine-tuned}   &	\textbf{0.951} &	\textbf{0.776} &	\textbf{7.091} &	29 & 9.52 \\
\botrule
\end{tabular}
\footnotetext[1]{ETT: Epoch training time in seconds.}
\end{minipage}
\end{center}
\end{table}

\begin{table}[ht]
\begin{center}
\begin{minipage}{\textwidth}
\caption{Multi-task Network Gesture Segmentation Results}\label{gesturesegmentationresults}%
\begin{tabular}{@{}llllllll@{}}
\toprule
                                                       & \textbf{Accuracy}     & \textbf{Edit} & \textbf{F1@0.1} & \textbf{F1@0.25} & \textbf{F1@0.5} & \textbf{ETT\textsuperscript{1}}\\
\midrule
\textbf{Sensors} \cite{goldbraikh_using_2022}            & 82.40 ± 6.58           & 85.99         & 89.09             & 85.80           & 71.41 & -   \\
\midrule
\multicolumn{7}{c}{\textbf{RGB Frontal View}} \\
\midrule

\textbf{Keypoints}          & 81.22 ± 5.61           & 83.61         & 86.39             & 82.62             & 67.41    & \textbf{9.21}     \\
\textbf{I3D}                & 83.11 ± 5.84           & \textbf{86.35}         & \textbf{89.10}             & \textbf{86.18}             & \textbf{73.76}   & 14.49     \\
\textbf{I3D + Keypoints}    & \textbf{83.24 ± 6.11}           & 85.60         & 88.33             & 84.90             & 71.85   & 14.75     \\
\midrule
\multicolumn{7}{c}{\textbf{RGB Closeup View}} \\

\midrule
\textbf{Keypoints}          & 84.16 ± 5.37          & 79.95       & 84.25           & 82.02          & 72.63    & \textbf{9.21}     \\
\textbf{I3D}                & 87.16 ± 4.72          & \textbf{84.66}       & \textbf{89.70}           & \textbf{88.33}          & \textbf{82.10 }    & 14.49    \\
\textbf{I3D + Keypoints}    & \textbf{87.69 ± 4.40}          & 83.32       & 88.16           & 86.72          & 79.99     & 14.75    \\
\midrule
\multicolumn{7}{c}{\textbf{RGB Multi-view}} \\
\midrule
\textbf{Keypoints}       & 85.39 ± 4.35       & 81.63           & 85.76           & 84.16           & 75.83     & \textbf{10.18}     \\
\textbf{I3D}             & 87.89 ± 4.13       & \textbf{85.76}           & \textbf{90.61}           & \textbf{89.08}           & \textbf{82.82}      & 20.29    \\
\textbf{I3D + Keypoints}             & \textbf{88.35 ± 4.15}       & 85.32           & 89.68          & 88.28           & 82.32      & 20.80    \\
\botrule
\end{tabular}
\footnotetext{The chosen MSTCN++ hyperparameters are learning rate = 0.001, number of feature maps = 64, prediction generation layers = 11, refinement layers = 10, refinement stages = 1.}
\footnotetext[1]{Epoch training time in seconds for MSTCN++.}
\end{minipage}
\end{center}
\end{table}

\section{Conclusion and Discussion}
\label{discussion}
When it comes to gesture segmentation, remote pose estimations showed comparable accuracy to sensors on the frontal view (81.81\% vs 82.40\%), and better accuracy on the close-up and mixed views (84.16\% and 85.39\%). Despite being less accurate than I3D features, they offer the advantage of context isolation, and support for concrete feedback through skill proxies. Furthermore, concatenating key points from multiple views leads to a 1.23\% increase in accuracy which leads us to believe that 3D pose estimations could outperform current results. Finally, we see that combining pose estimations and I3D features results in up to 0.53\% added accuracy.

As for skill assessment and training feedback, the proxies presented in section \ref{methods:5} highlight the notable distinctions between novices and experts. This allows the system to automatically offer task-specific feedback, such as instructing the user to keep the suture away from the tissue while cutting or to straighten the index finger while holding the needle driver. This is accomplished by comparing the proxy values of new samples to the average of the experts during the specific task, and providing feedback if the difference surpasses a pre-determined threshold that is independently adjusted for each proxy. This is a significant improvement over prior methods such as those used by Liu et al. \cite{liu_towards_2021}, who relied on semantic visual features and tool trajectory, and Goldbraikh et al. \cite{goldbraikh_video-based_2022}, who used global metrics.

An important limitation to note is our use of 2D pose estimations instead of 3D. This limits some of our proxies such as hand orientation \ref{methods-hand-orientation:5.2} to a single camera angle. Given that we used the videos of the frontal view in our dataset, applying this proxy was successful. When applying it using different camera angles, a 3D pose would provide a more accurate result.

To conclude, this research aims to bridge the gap in the application of novel computer vision algorithms to the domain of surgical training and monitoring. We demonstrate how 2D pose estimation can be applied to new open surgery datasets, and how it can be utilized for gesture segmentation and skill assessment. By creating the proxies depicted in section \ref{methods:5}, and assessing known groups' validity evidence with our dataset, we demonstrate how the expert’s advice can be automated through the surgical skill proxy methodology. This paves the way to work towards a fully automated surgical training framework requiring only a performance video.

\section*{Declarations}

\textbf{Acknowledgments} This research was partially funded by the Technion Center for Machine Learning and Intelligent Systems and by Israel PBC-VATAT.\\
\textbf{Conflict of interest} The authors declare that they have no conflict of interest. \\
\textbf{Ethical approval} All procedures performed in studies involving human participants were in accordance with the ethical standards of the institutional and/or national research committee and with the 1964 Helsinki Declaration and its later amendments or comparable ethical standards. \\
\textbf{Informed consent} Informed consent was obtained from all individual participants included in the study. \\
\textbf{Code availability} The code will be released after publication at https://github.com/edybk/Hand-Pose-Estimation-For-Surgical-Training

\bibliography{references}

\end{document}